\documentclass[conference]{IEEEtran}
\usepackage[utf8]{inputenc}
\usepackage{graphicx}
\usepackage{amsmath,amssymb}
\usepackage{cite}
\usepackage{url}
\usepackage{multirow}
\usepackage{caption}
\usepackage{subcaption}
\usepackage{hyperref}
\usepackage{algorithm}
\usepackage{algpseudocode}
\usepackage{array} 
\usepackage{booktabs}
\usepackage[utf8]{inputenc}
\usepackage{placeins}
\usepackage{cleveref}
\newcommand{\degree}{\ensuremath{^\circ}}

\title{
AI-driven Web Application for Early Detection of Sudden Death Syndrome (SDS) in Soybean Leaves Using Hyperspectral Images and Genetic Algorithm
}

\author{
    Pappu Kumar Yadav$^{1*}$, 
    Rishik Aggarwal$^{1}$, 
    Supriya Paudel$^{1}$, 
    Amee Parmar$^{1}$, 
    Hasan Mirzakhaninafchi$^{1}$, \\
    Zain Ul Abideen Usmani$^{1}$, 
    Dhe Yeong Tchalla$^{1}$, 
    Shyam Solanki$^{2}$,
    Ravi Mural$^{2}$,
    Sachin Sharma$^{2}$\\
    Thomas F. Burks$^{3}$, 
    Jianwei Qin$^{4}$, 
    Moon S. Kim$^{4}$ \\
    \\
    $^1$Machine Vision and Optical Sensor (MVOS) Lab, Dept. of Agricultural and Biosystems Engineering,\\
    South Dakota State University, Brookings, SD, USA \\
    $^2$Department of Agronomy, Horticulture and Plant Science, South Dakota State University, Brookings, SD, USA \\
    $^3$Department of Agricultural and Biological Engineering, University of Florida, Gainesville, FL, USA \\
    $^4$USDA-ARS Environmental Microbial and Food Safety Lab, Beltsville Agricultural Research Center, MD, USA \\
    \\
    *Corresponding author: \texttt{pappu.yadav@sdstate.edu}
}

\begin{document}
\maketitle
\begin{abstract}
Sudden Death Syndrome (SDS), caused by \textit{Fusarium virguliforme}, poses a significant threat to soybean production. This study presents an AI-driven web application for early detection of SDS on soybean leaves using hyperspectral imaging, enabling diagnosis prior to visible symptom onset. Leaf samples from healthy and inoculated plants were scanned using a portable hyperspectral imaging system (398--1011~nm), and a Genetic Algorithm was employed to select five informative wavelengths (505.4, 563.7, 712.2, 812.9, and 908.4~nm) critical for discriminating infection status. These selected bands were fed into a lightweight Convolutional Neural Network (CNN) to extract spatial--spectral features, which were subsequently classified using ten classical machine learning models. Ensemble classifiers (Random Forest, AdaBoost), Linear SVM, and Neural Net achieved the highest accuracy ($>$98\%) and minimal error across all folds, as confirmed by confusion matrices and cross-validation metrics. Poor performance by Gaussian Process and QDA highlighted their unsuitability for this dataset. The trained models were deployed within a web application that enables users to upload hyperspectral leaf images, visualize spectral profiles, and receive real-time classification results. This system supports rapid and accessible plant disease diagnostics, contributing to precision agriculture practices. Future work will expand the training dataset to encompass diverse genotypes, field conditions, and disease stages, and will extend the system for multiclass disease classification and broader crop applicability.
\end{abstract}

\vspace{1em}
\noindent\textbf{Code and Web App:}\\
\noindent GitHub:\href{https://github.com/MVOSlab-sdstate/soybean-SDS-leaf-classifier}{https://github.com/MVOSlab-sdstate/soybean-SDS-leaf-classifier} \\
Web App: \href{https://soybeansdsclassifiermvoslabsdsu.streamlit.app/}{https://soybeansdsclassifiermvoslabsdsu.streamlit.app/}

\begin{IEEEkeywords}
Hyperspectral imaging, Sudden Death Syndrome, Soybean, Genetic algorithm, Web application, Deep learning, Plant disease detection.
\end{IEEEkeywords}

\section{Introduction}

Since its first report in Arkansas in 1971, Sudden Death Syndrome (SDS) has spread across all major soybean (\textit{Glycine max} L.) producing regions of the United States, including South Dakota~\cite{herrmann2018fusarium}. Economic impacts have been severe as between 1996 and 2007, SDS cost growers roughly \$100 million annually~\cite{herrmann2018fusarium, wrather2009effects}, and in 2022 the disease was estimated to reduce U.S.\ production by more than 18 million bushels~\cite{cpn2022losses}. Under favorable conditions for the causal pathogen, \textit{Fusarium virguliforme}, yield losses of 50–80\% are common, and total crop failure has been documented in extreme cases~\cite{herrmann2018fusarium, cui2014method}.

Foliar symptoms usually appear after flowering and intensify during pod fill. Early signs such as small, pale green spots that progress to interveinal chlorosis and necrosis can be confused with brown stem rot (BSR) and stem canker. Together with white mold, SDS now ranks among the most damaging diseases in the north-central United States, threatening 40–50 million acres of soybean each year~\cite{bradley2021soybean}. Precise, timely diagnosis is therefore critical.

Conventional detection relies on field scouting followed by laboratory confirmation (microscopy or PCR)~\cite{li2003molecular, cho2001comparison}. Although PCR is sensitive and specific, shipping samples and processing in diagnostic labs generally requires five to ten business days, long enough for the disease to progress beyond manageable levels.

Hyperspectral imaging (HSI) combined with artificial-intelligence (AI) analytics offers a non-invasive route to pre-symptomatic disease detection as the narrow spectral bands capture subtle biochemical changes before visible lesions form~\cite{frederick2023bands, yadav2022citrus, yadav2023spiea}. However, high spectral dimensionality often introduces redundancy and computational inefficiency, making feature selection a critical step in building accurate models. Genetic Algorithms (GAs), inspired by the process of natural selection, have been successfully applied in plant disease classification tasks to identify optimal combinations of spectral bands that preserve class separability while reducing model complexity~\cite{sharma2016ga, yin2019hsiband, frederick2023bands}. These evolutionary optimization techniques are particularly well suited for hyperspectral data, where exhaustive search becomes computationally prohibitive due to the large number of band combinations.

To address these challenges and bridge the gap between research and real-world use, we employed a portable hyperspectral camera that records 348 bands in the 398–1011 nm range and coupled it with a deep learning pipeline that uses GA to select five informative bands. A Convolutional Neural Network (CNN) then extracts spectral–spatial features from these bands for binary classification of soybean leaves as either healthy (H) or SDS-infected (I). The entire system is integrated into a browser based web application, allowing users to upload hyperspectral image cubes and receive classification results and spectral plots in an average time of under two minutes per leaf sample.

This study was focused on four specific objectives: (i) acquire a curated hyperspectral dataset of healthy and SDS-infected soybean leaves throughout the growing season (V3, V4, V5 and V6 growth stages); (ii) apply a Genetic Algorithm to select the five most informative bands for SDS classification; (iii) design a CNN-based spatial feature extractor and train ten classical machine learning classifiers on the selected bands for healthy (H) versus SDS-infected (I) prediction; and (iv) develop a browser based web interface that enables users to upload soybean leaf hypercubes, receive classification results, and visualize central-pixel spectra almost instantaneously.
\section{Material and Methods}
\subsection{Experiment Setup}

To generate hyperspectral data for early stage classification of soybean leaves as either healthy or infected with Sudden Death Syndrome (SDS), a controlled greenhouse experiment was conducted at South Dakota State University, Brookings, SD.

\textit{Fusarium virguliforme} isolate SLFV\_1, originally collected from naturally infected soybean roots in South Dakota fields, was used for inoculum preparation in this study. Root samples were washed under running tap water to remove soil debris, surface sterilized using 1.5\% sodium hypochlorite solution for 2 minutes, and rinsed three times with sterile distilled water~\cite{li2003molecular, rupe1989fusarium}. Small symptomatic root segments were plated on potato dextrose agar (PDA; VWR) and incubated at 25\,\degree C for 5–7 days. Colonies were tentatively identified as \textit{F.~virguliforme} based on morphological features, including pigmentation and conidial structure, and identity was confirmed using PCR with species-specific primers targeting the translation elongation factor 1-$\alpha$ gene~\cite{li2003molecular}.

The SLFV\_1 isolate was stored at --20\,\degree C and later revived on sterilized PDA medium in culturing-grade Petri dishes (BD Biosciences). Actively growing fungal mycelia were transferred as 5\,mm agar plugs to overnight-soaked sorghum grains, which had been autoclaved in 2-liter glass flasks (VWR) 24 hours prior. The inoculated sorghum was incubated at 27\,\degree C for 15–20 days with gentle mixing every other day to promote uniform colonization. After confirming adequate fungal growth, the colonized grain was dried for 48 hours and stored at 4\,\degree C until use as SDS inoculum~\cite{chawla2013public, herman2023evaluation, okello2023fusarium}. Greenhouse inoculations were conducted using the layered inoculum method at planting, with inoculated sorghum applied in a uniform layer below the seed, as previously described for SDS and other \textit{Fusarium} spp.~\cite{chawla2013public, herman2023evaluation, okello2023fusarium}.

\begin{figure}[htbp]
    \centering
    \includegraphics[width=0.48\textwidth]{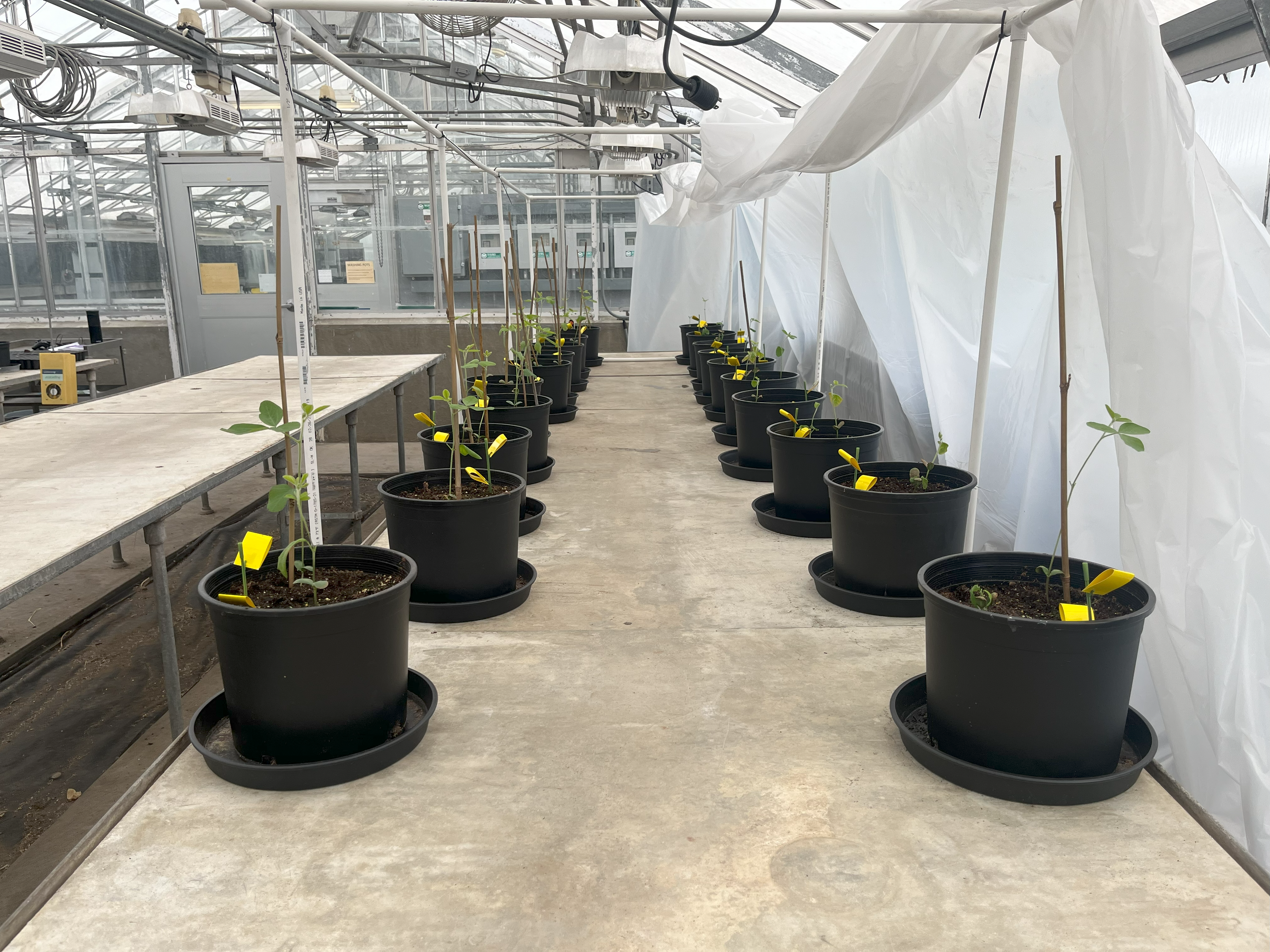}
    \caption{Greenhouse setup showing soybean plants grown in pots (three plants per pot). Sticky yellow tags were placed for pest monitoring. The left row consists of plants inoculated with \textit{Fusarium virguliforme}, while the right row contains uninoculated healthy control plants.}
    \label{fig:greenhouse_setup}
\end{figure}

For the greenhouse experiment, a total of 60 soybean (\textit{Glycine max} L.) plants of variety AP0720 (Ag Performance, Buffalo Center, IA, USA) were grown in 20 pots (three plants per pot) (Figure~\ref{fig:greenhouse_setup}). Ten pots (30 plants) were inoculated with fresh \textit{F. virguliforme} culture using a grain sorghum inoculum method, where autoclaved sorghum grains were colonized by the fungus and mixed into the potting medium~\cite{rupe1989fusarium}. The remaining ten pots (30 plants) were left uninoculated to serve as healthy controls.

Soybean vegetative growth stages were determined using the Iowa State University Extension criteria, which define vegetative (V) stages based on the number of fully developed trifoliolate leaves~\cite{iastate_growth}. A growth stage is considered to have occurred when at least 50\% of the plants exhibit the defined leaf development. For example, the V3 stage corresponds to three unfolded trifoliolate leaves above the unifoliolate node. Using this method, plants were staged, and leaf samples were collected from both inoculated and control pots at the V3, V4, V5, and V6 stages, irrespective of visible symptom expression.

Although hyperspectral data were acquired across all four stages, this paper focuses specifically on results obtained from the V3 stage. This early stage was selected due to its relevance for timely disease intervention and the observed consistency in spectral response across replicates.

\subsection{Hyperspectral Imaging System (HSI)}

\begin{figure}[htbp]
    \centering
    \includegraphics[width=0.48\textwidth]{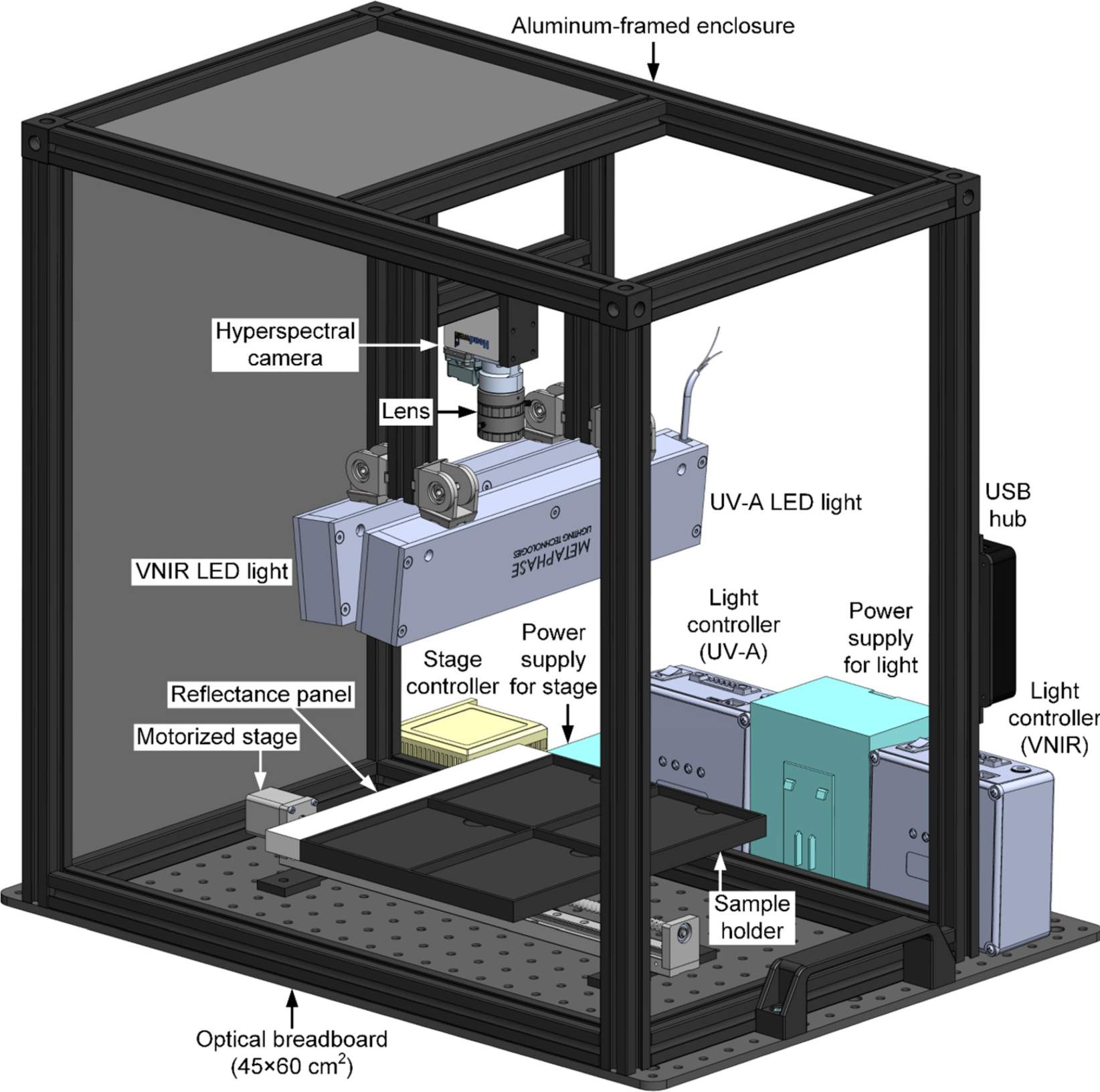}
    \caption{A portable HyperSpectral Imaging (HSI) system that was used for this study. ~\cite{frederick2025supervised}}
    \label{fig:hsisystem}
\end{figure}

Hyperspectral data were acquired with a portable line–scan system (Figure~\ref{fig:hsisystem}) originally developed at the USDA-ARS Environmental Microbial and Food Safety Laboratory and recently updated for plant disease studies~\cite{yadav2022citrus}.  The instrument records 348 contiguous bands spanning 398–1011\,nm.  Illumination is supplied by two linear LED arrays: a broadband VNIR bank (428, 650, 810, 850, 890, 910, and 940\,nm) for reflectance imaging and a 365\,nm UV-A bank for fluorescence excitation.  LED intensities are software-dimmable, and the bars are mounted at a 6$^{\circ}$ angle to ensure uniform line lighting across the sample tray.

Spectral data are captured with a compact VNIR Nano-Hyperspec camera (Headwall Photonics, Bolton, MA, USA) equipped with a 12-bit CMOS detector (1936\,$\times$\,1216\,px) and a 5\,mm lens.  A $>$400\,nm long-pass filter blocks second-order UV light.  The camera is fixed 285\,mm above a motorised translation stage (FUYU Technology, Chengdu, China) that moves the sample under the field of view, yielding 0.33\,mm\,px$^{-1}$ spatial resolution over an 810\,px line.  A Labsphere white tile and a black sample tray are scanned in tandem for reflectance normalisation.  All optical components are enclosed in a 56\,$\times$\,45\,$\times$\,60\,cm aluminium housing to eliminate ambient light.

System control and realtime visualisation were implemented in LabVIEW 2022.  LED drivers communicate via UDP, the camera via USB, and the stage via RS-232.  Each scan generates a band-interleaved-by-line (BIL) cube for reflectance and a separate cube for fluorescence; only the reflectance data were analysed here.  This platform has proven effective for other plant pathogen and quality assessment tasks, including citrus peel defect and leaf disease detection~\cite{frederick2023bands}.

The pseudocode describing the process of spatial and spectral binning techniques that are applied after scanning the leaf samples using the HSI system is given in Algorithm 1.

\begin{algorithm}[htbp]
\caption{HSI Pre-processing: Flat-Field Correction, Spectral and Spatial Binning}
\label{alg:preproc}
\begin{algorithmic}[1]
\Require Raw hyperspectral cube $H_{\text{raw}}\!\in\!\mathbb{R}^{250 \times 810 \times 348}$
\Require White reference $W$ and dark reference $D$ (same spatial–spectral size)
\Ensure Calibrated and binned cube $H' \in \mathbb{R}^{250 \times 270 \times 116}$

\Statex \textbf{Flat-field calibration}
\State $H_{\text{ref}} \gets \displaystyle \frac{H_{\text{raw}} - D}{W - D}$ \Comment{Convert raw DN to reflectance}

\Statex \textbf{Spectral binning (348 $\rightarrow$ 116)}
\State Choose bin size $k_s = 3$ (three adjacent bands per bin)
\For{$b = 1$ to $348$ \textbf{step} $k_s$}
   \State Compute \emph{mean} of bands $b \dots b+k_s-1$  \Comment{Other aggregations also possible}
\EndFor
\State $H_{\text{spec}} \gets$ stack of 116 mean–band images  \Comment{$250 \times 810 \times 116$}

\Statex \textbf{Spatial trim / binning (810 $\rightarrow$ 270)}
\State Select central region of interest or downsample by factor $k_p=3$
\For{$i = 1$ to $116$}
   \State $H_{\text{spat}}[:, :, i] \gets$ bicubic\,/\,area resize of $H_{\text{spec}}[:, :, i]$ to $250 \times 270$
\EndFor

\Statex \textbf{Return}
\State $H' \gets H_{\text{spat}}$ \Comment{Final cube $250 \times 270 \times 116$}
\end{algorithmic}
\end{algorithm}

\subsection{Genetic Algorithm for Band Selection}
\label{subsec:GA}

To reduce data dimensionality and isolate the most informative spectral bands for classifying soybean leaves as healthy or infected with Sudden Death Syndrome (SDS), we implemented a Genetic Algorithm (GA)-based spectral feature selection approach. GA is a population based evolutionary optimization technique inspired by natural selection, where a population of candidate solutions called chromosomes evolves over successive generations through crossover, mutation, and selection operations~\cite{goldberg1989genetic, nagasubramanian2018hyperspectral}. In our implementation, each chromosome was encoded as an ordered list of five unique integers, each representing an index among the 116 available spectral bands extracted after spectral binning. The initial population consisted of 12 randomly generated chromosomes. Evolution was carried out over eight generations, with two point crossover applied at a probability of 0.7 and uniform integer mutation at 0.2. Tournament selection (with tournament size 3) was used to select individuals for the next generation. The fitness of each chromosome was evaluated as the classification accuracy of a lightweight convolutional neural network (CNN), trained on the reflectance data corresponding to the five bands selected by that chromosome. Chromosomes containing invalid or duplicate indices were excluded from evaluation. The entire GA process was implemented using the \texttt{DEAP} library in Python ~\cite{DEAP_JMLR2012}, and multiprocessing was employed to parallelize the CNN-based fitness evaluations. After the final generation, the chromosome with the highest validation accuracy was chosen as the optimal set of bands. This five band subset was used in all further analysis.

\subsection{Convolutional Neural Network Architecture, Training and Evaluation}
\label{subsec:CNN}

After GA had reduced the hyperspectral cube to five informative bands, each soybean leaf image was a tensor of size $125 \times 100 \times 5$ (rows, columns, bands). The original 348-band hyperspectral images were first flat-field calibrated and spectrally binned to generate 116 bands, of which the first 6 and last 9 were excluded due to sensor-related artifacts, yielding 101 usable bands. A compact convolutional neural network (CNN) was devised to distill these 3-D inputs into a low-dimensional representation suitable for binary classification (healthy vs.\ SDS-infected). The network consists of a single $3 \times 3$ convolutional layer with 32 filters and ReLU activation, followed by a $2 \times 2$ max-pooling layer, a flattening operation, one fully connected layer of 64 ReLU units and a two-node \textit{softmax} output. The model is trained fold-wise with the Adam optimiser ($\alpha = 10^{-3}$) and \texttt{categorical\_crossentropy} loss for up to 50 epochs (batch = 16), employing early stopping on validation loss. Five-fold stratified cross-validation provides an unbiased estimate of performance. In each fold the best CNN (lowest validation loss) serves as a feature extractor whose 64-dimensional activations feed ten classical classifiers: KNN, linear and RBF SVM, Gaussian Process, Decision Tree, Random Forest, MLP, AdaBoost, Naïve Bayes and QDA.

\begin{figure*}[t]
    \centering
    \includegraphics[width=\textwidth]{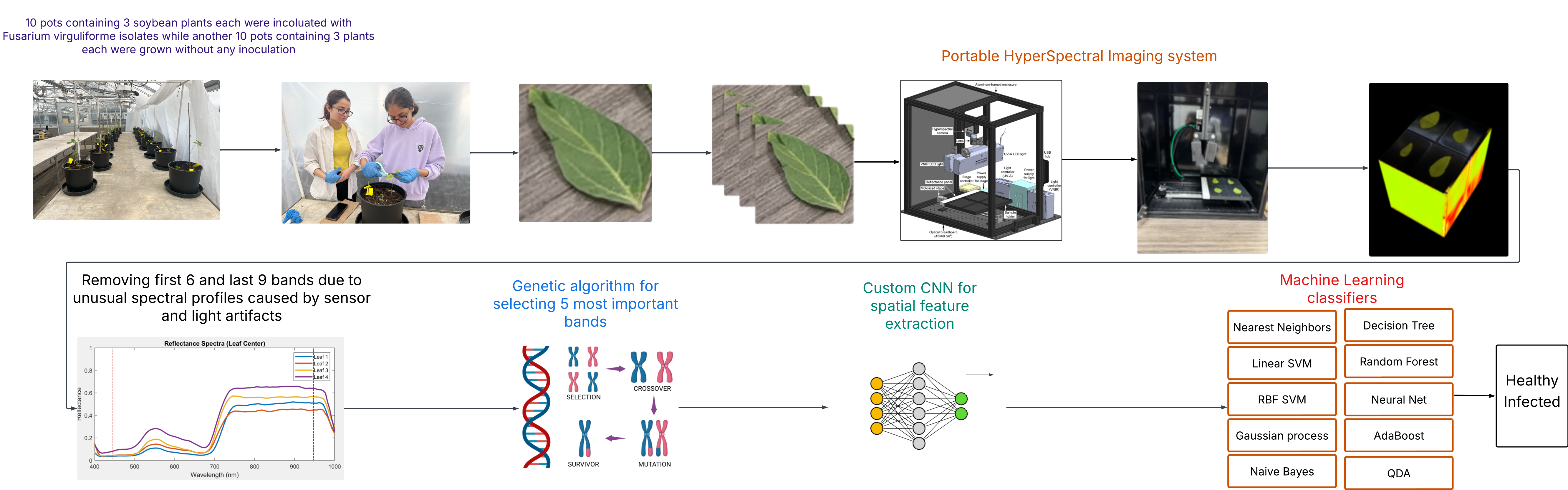}
    \caption{Workflow pipeline showing soybean SDS classification: greenhouse inoculation, leaf sampling, hyperspectral image acquisition, removal of sensor-affected bands, genetic algorithm-based band selection, CNN-based spatial feature extraction, and final classification using multiple machine learning classifiers.}
    \label{fig:pipeline}
\end{figure*}

The trained models were assessed with five performance metrics. The Overall accuracy is given by equation:
\begin{equation}
\text{Accuracy} \;=\; \frac{\text{TP}+\text{TN}}{\text{TP}+\text{FP}+\text{TN}+\text{FN}},
\end{equation}
where, TP, TN, FP and FN represent true postive, true negative, false positive and false negative respectively; while mean-squared error (MSE) for misclassification is given by equation:
\begin{equation}
\text{MSE} = \frac{1}{n}\sum_{i=1}^{n}\bigl(y_i - \hat{y}_i\bigr)^2,
\end{equation}
where $y_i$ and $\hat{y}_i$ are the true and predicted class labels for the $i^{\text{th}}$ sample ($n$ samples per fold). Precision and recall quantify class-specific performance and are given by equations:
\begin{equation}
\text{Precision} = \frac{\text{TP}}{\text{TP}+\text{FP}}, 
\end{equation}
\begin{equation}
\text{Recall}    = \frac{\text{TP}}{\text{TP}+\text{FN}},
\end{equation}
while their harmonic mean is used to calculate the F1-score which is given by the equation:
\begin{equation}
\text{F1} = 2\,\frac{\text{Precision}\times\text{Recall}}{\text{Precision}+\text{Recall}}.
\end{equation}
All metrics are averaged over the five folds to give robust, generalizable estimates suitable for real-time deployment within a web application interface. The entire workflow is shown in (Figure~\ref{fig:pipeline}).

\subsection{Streamlit-based Web Application}
To make the trained SDS‐classification pipeline readily available to growers, extension specialists, plant pathologists and researchers, we implemented a lightweight graphical front-end using the \texttt{Streamlit} framework.\footnote{\url{https://streamlit.io}} ~\cite{streamlit2023}. The complete source code is archived on GitHub at  
\url{https://github.com/MVOSlab-sdstate/soybean-SDS-leaf-classifier} and a live instance is hosted at  
\url{https://soybeansdsclassifiermvoslabsdsu.streamlit.app/}.  

The app accepts one or more hyperspectral leaf cubes saved in \texttt{.mat} format. After upload, it (i) removes the first six and last nine bands (sensor artefacts) to obtain a \(125\times100\times101\) reflectance cube, (ii) displays an RGB rendering using bands 92, 83, and 54, (iii) plots the central-pixel spectrum for quick visual inspection, (iv) feeds the five GA-selected bands to a frozen CNN feature extractor (\texttt{cnn\_fold\_5.keras}), and (v) submits the resulting features to one of ten pre-trained classical classifiers selected by the user (e.g., Random Forest, RBF-SVM, KNN).

All model files are stored on Google Drive and fetched on demand using \texttt{gdown}, keeping the Streamlit deployment lightweight while enabling model updates without redeployment. For each sample, the predicted label (``Healthy'' or ``Infected (SDS)'') and its spectral plot are shown directly in the browser. Users can download a professionally formatted PDF containing all results, spectra, timestamps, and SDSU branding; this report is compiled on the fly using \texttt{ReportLab}. Styling that includes a darkened soybean field background, institutional logo, and semi-transparent input widgets is handled entirely via embedded Cascading Style Sheets (CSS)~\cite{lie2005css, wolf2017css}, minimizing external dependencies.

Image and spectral inputs are processed on the backend using optimized \texttt{NumPy} and \texttt{OpenCV} pipelines to ensure sub-second latency for small to medium sized datasets. The user interface adapts seamlessly to both desktop and mobile resolutions, making it suitable for use in the field or during extension events.

\begin{figure*}[t]
    \centering
    \includegraphics[width=\textwidth]{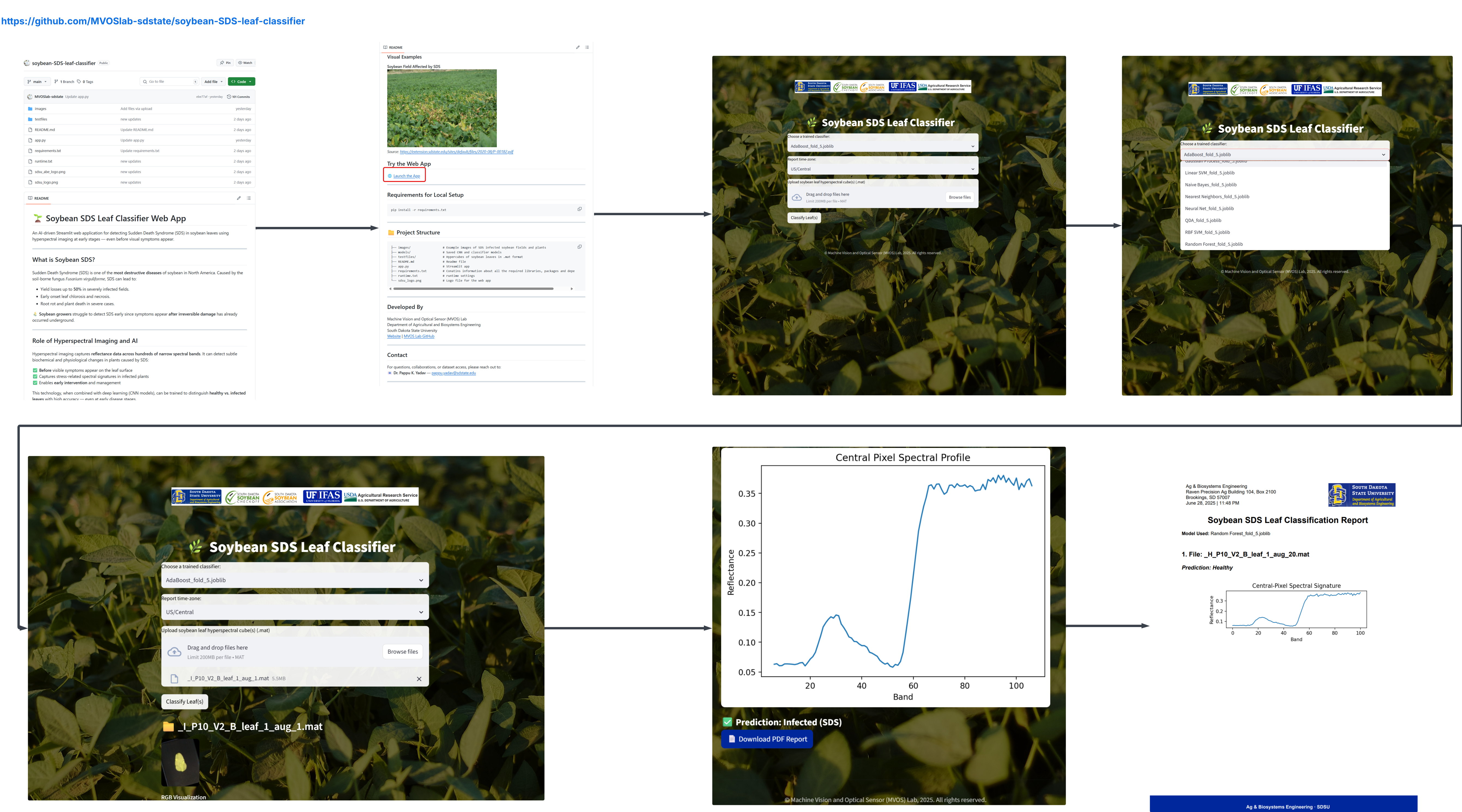}
    \caption{Flowchart describing steps on the usage of the web app for soybean SDS disease detection using hyperspectral images and trained machine learning classifiers.}
    \label{fig:webapp}
\end{figure*}

Because the application logic is fully containerized, it can be run locally, on institutional servers, or on any cloud service that supports Python~3.10 and Streamlit~1.30+. The steps guiding on the application of the web app is described in Figure~\ref{fig:webapp}.

\section{Results and Discussion}\label{sec:results}

\begin{figure*}[t]
    \centering
    % ---------- first row ----------
    \subfloat[Nearest Neighb.]{\includegraphics[width=.19\linewidth]{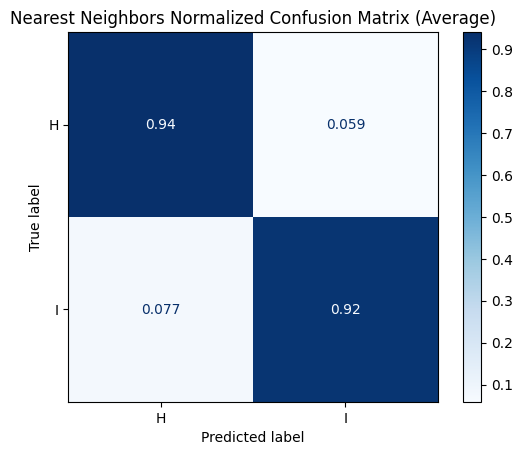}}\hfill
    \subfloat[Linear SVM]{\includegraphics[width=.19\linewidth]{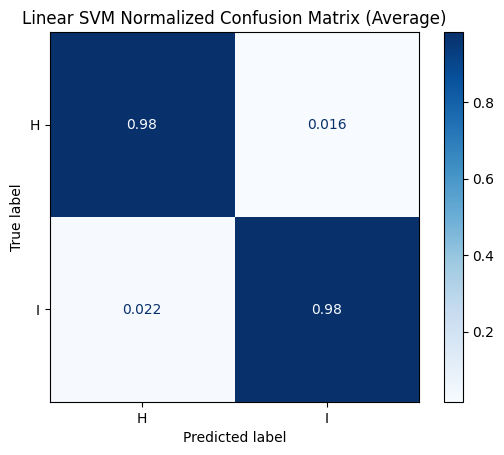}}\hfill
    \subfloat[RBF SVM]{\includegraphics[width=.19\linewidth]{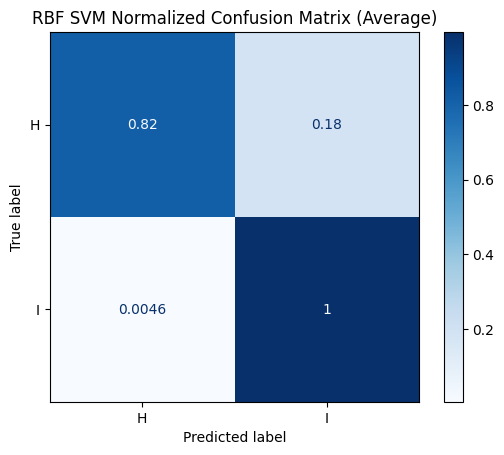}}\hfill
    \subfloat[Gauss.~Process]{\includegraphics[width=.19\linewidth]{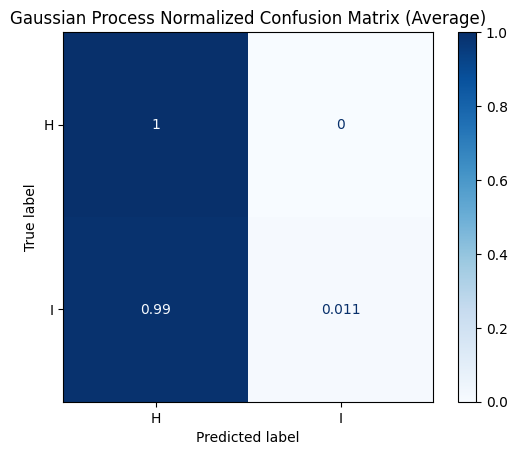}}\hfill
    \subfloat[Decision Tree]{\includegraphics[width=.19\linewidth]{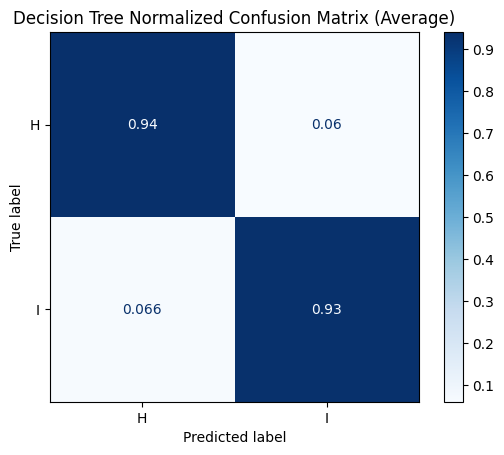}}\\[4pt]
    % ---------- second row ----------
    \subfloat[Random Forest]{\includegraphics[width=.19\linewidth]{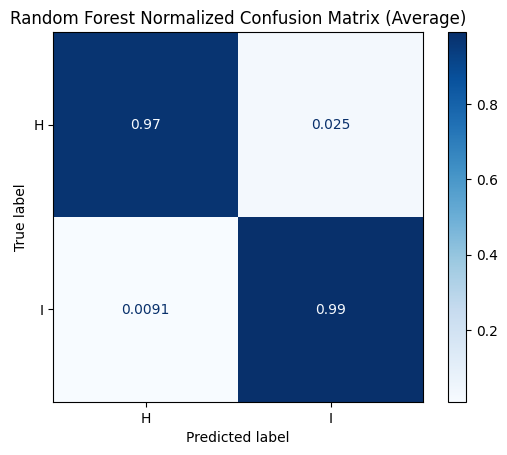}}\hfill
    \subfloat[Neural Net]{\includegraphics[width=.19\linewidth]{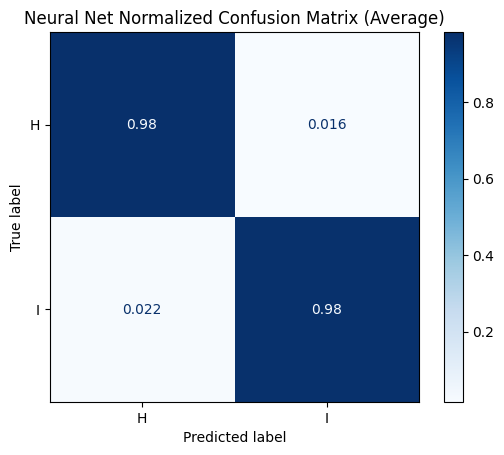}}\hfill
    \subfloat[AdaBoost]{\includegraphics[width=.19\linewidth]{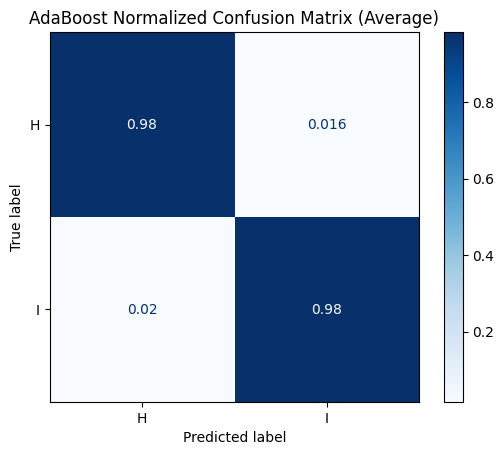}}\hfill
    \subfloat[Naive Bayes]{\includegraphics[width=.19\linewidth]{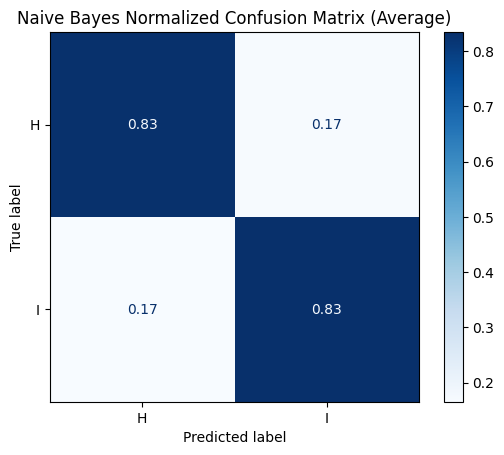}}\hfill
    \subfloat[QDA]{\includegraphics[width=.19\linewidth]{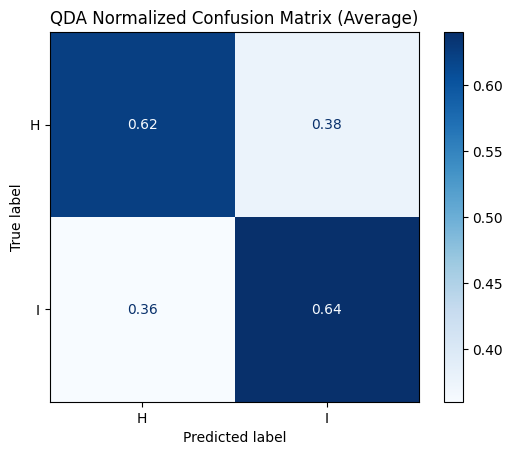}}
    \caption{Five-fold mean, class-normalised confusion matrices for the ten classical
             classifiers evaluated. \textbf{H} = healthy leaf; \textbf{I} = \textit{F. virguliforme}–infected
             leaf.  Darker cells along the main diagonal indicate higher correct-classification
             rates.}
    \label{fig:cm_grid}
\end{figure*}

Table~\ref{tab:ga_selected_bands} lists the five hyperspectral bands selected by the Genetic Algorithm (GA) as the most informative for classifying soybean leaf samples into healthy and SDS-infected categories. These bands, centered at 505.4, 563.7, 712.2, 812.9, and 908.4\,nm, span both the visible and near-infrared (NIR) regions of the spectrum. The presence of bands in the NIR region aligns with previous findings that stress-related changes in leaf structure and water content are more pronounced beyond 700\,nm. The combination of visible and NIR features suggests that the GA effectively identified spectral regions sensitive to both pigment-level and structural changes associated with SDS infection.

\begin{table}[t]
\centering
\caption{Bands selected by the Genetic Algorithm and their corresponding wavelengths.}
\label{tab:ga_selected_bands}
\begin{tabular}{lc}
\toprule
\textbf{Band Index} & \textbf{Wavelength (nm)} \\
\midrule
21 & 505.4 \\
32 & 563.7 \\
60 & 712.2 \\
79 & 812.9 \\
97 & 908.4 \\
\bottomrule
\end{tabular}
\end{table}

\begin{figure}[H]
    \centering
    \includegraphics[width=0.4\textwidth, height=5 cm]{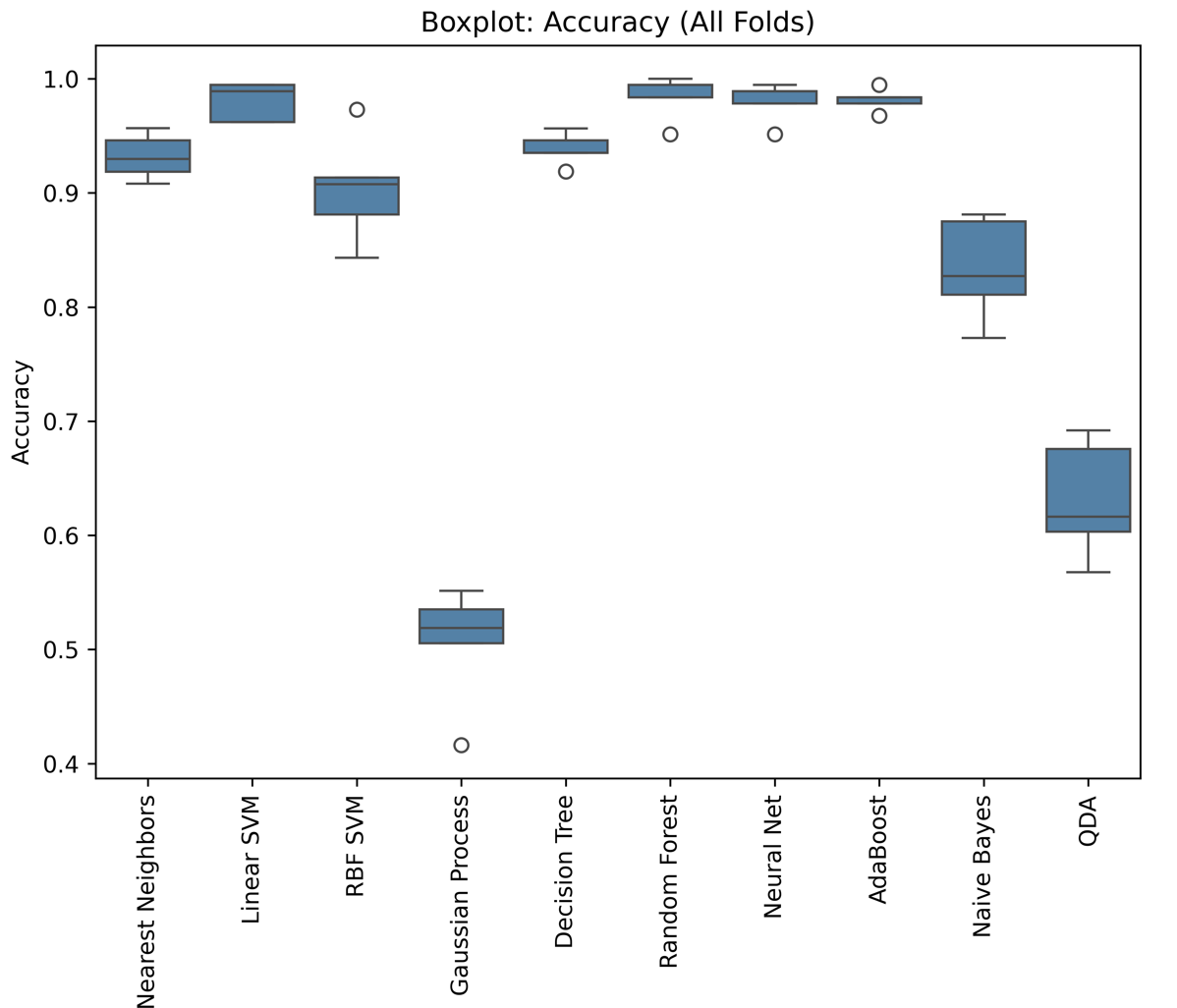}
    \caption{Box plots showing mean accuracies across five folds for each of the ten classifiers.}
    \label{fig:acc_boxplot}
\end{figure}

\begin{figure}[H]
    \centering
    \includegraphics[width=0.4\textwidth, height=5 cm]{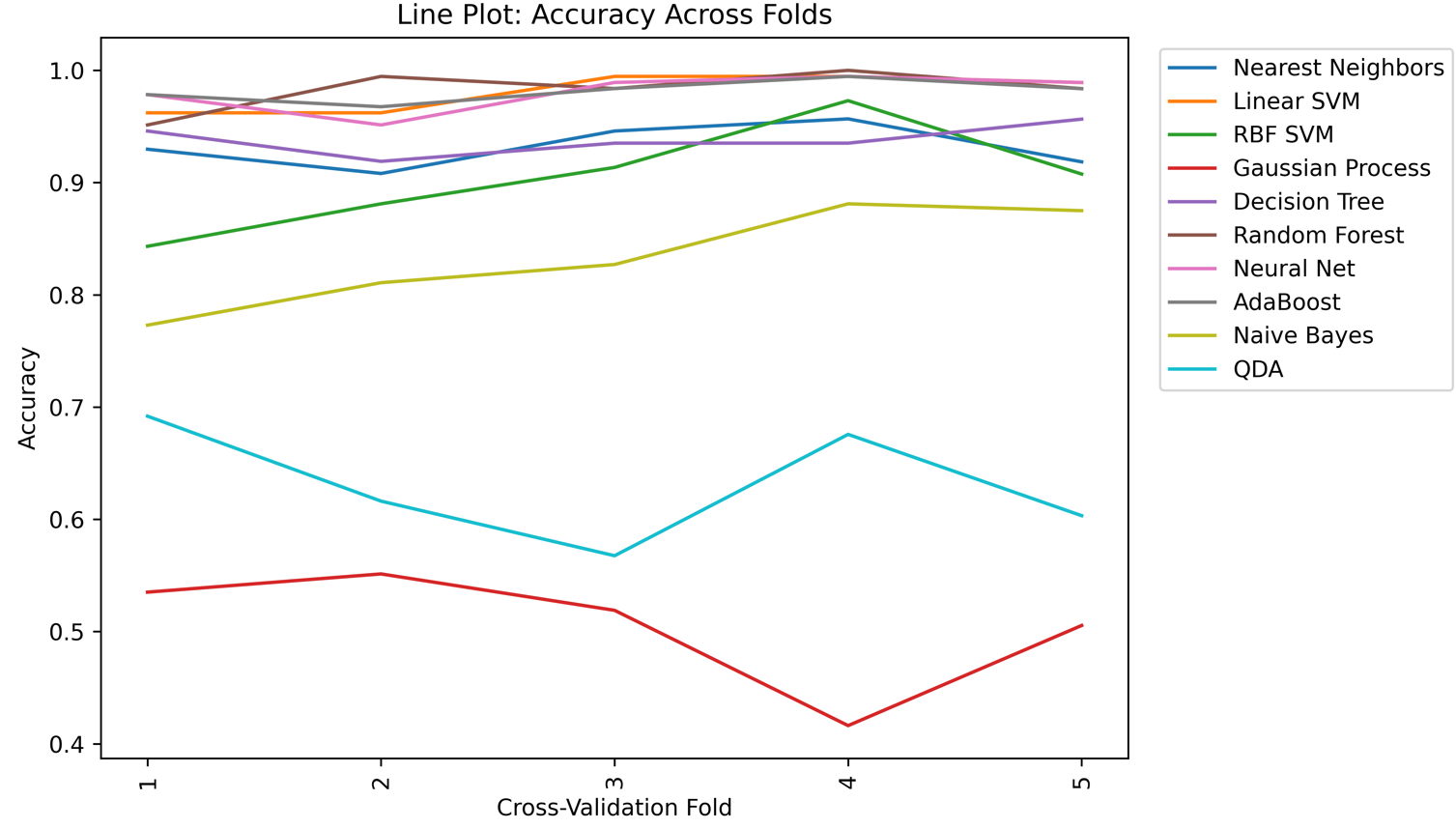}
    \caption{Line plots showing accuracies across five folds for each of the ten classifiers.}
    \label{fig:acc_folds}
\end{figure}

\begin{figure}[H]
    \centering
    \includegraphics[width=0.4\textwidth, height=5 cm]{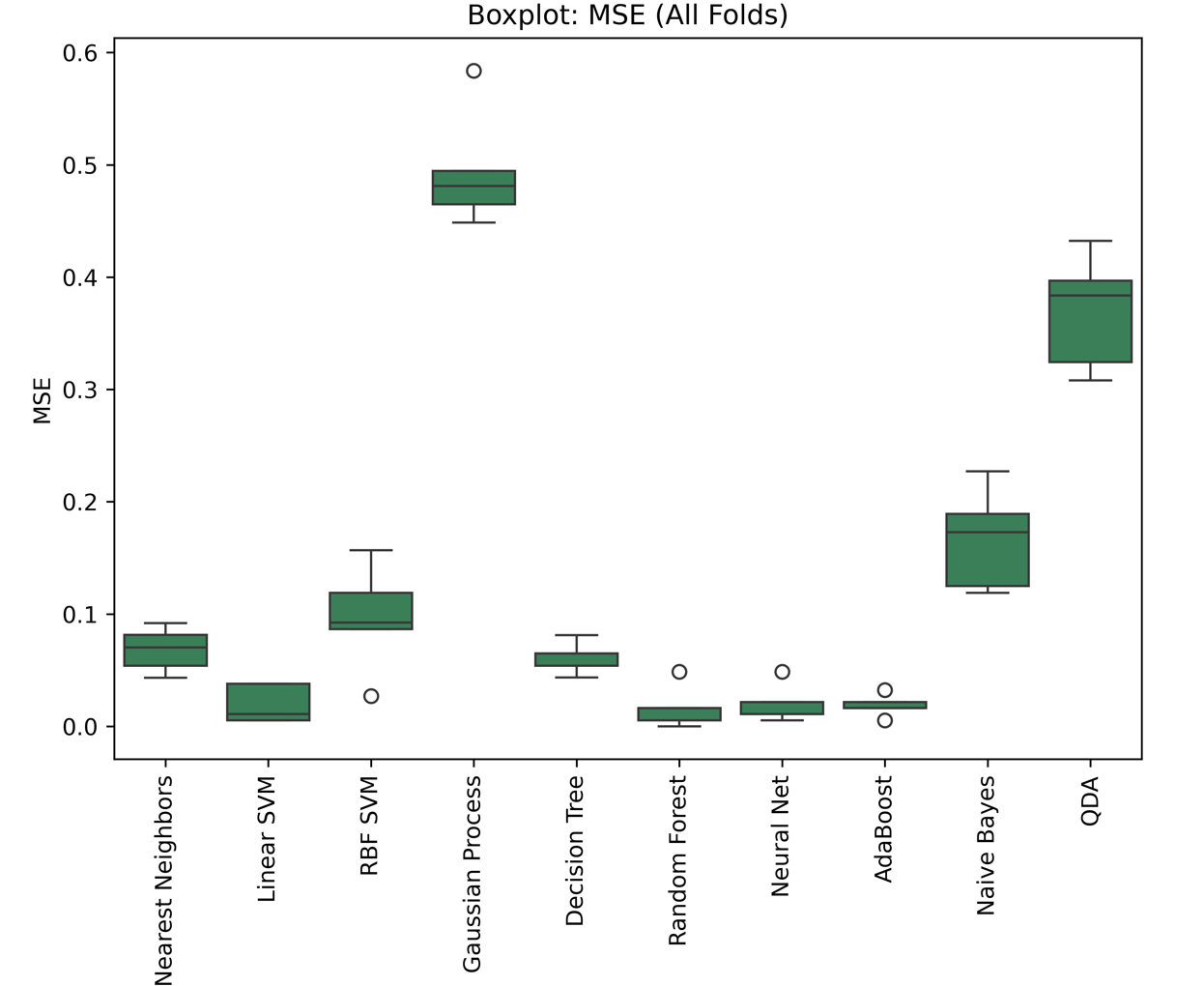}
    \caption{Box plots showing mean square error across five folds for each of the ten classifiers.}
    \label{fig:mse_boxplot}
\end{figure}

\begin{figure}[H]
    \centering
    \includegraphics[width=0.4\textwidth, height=5 cm]{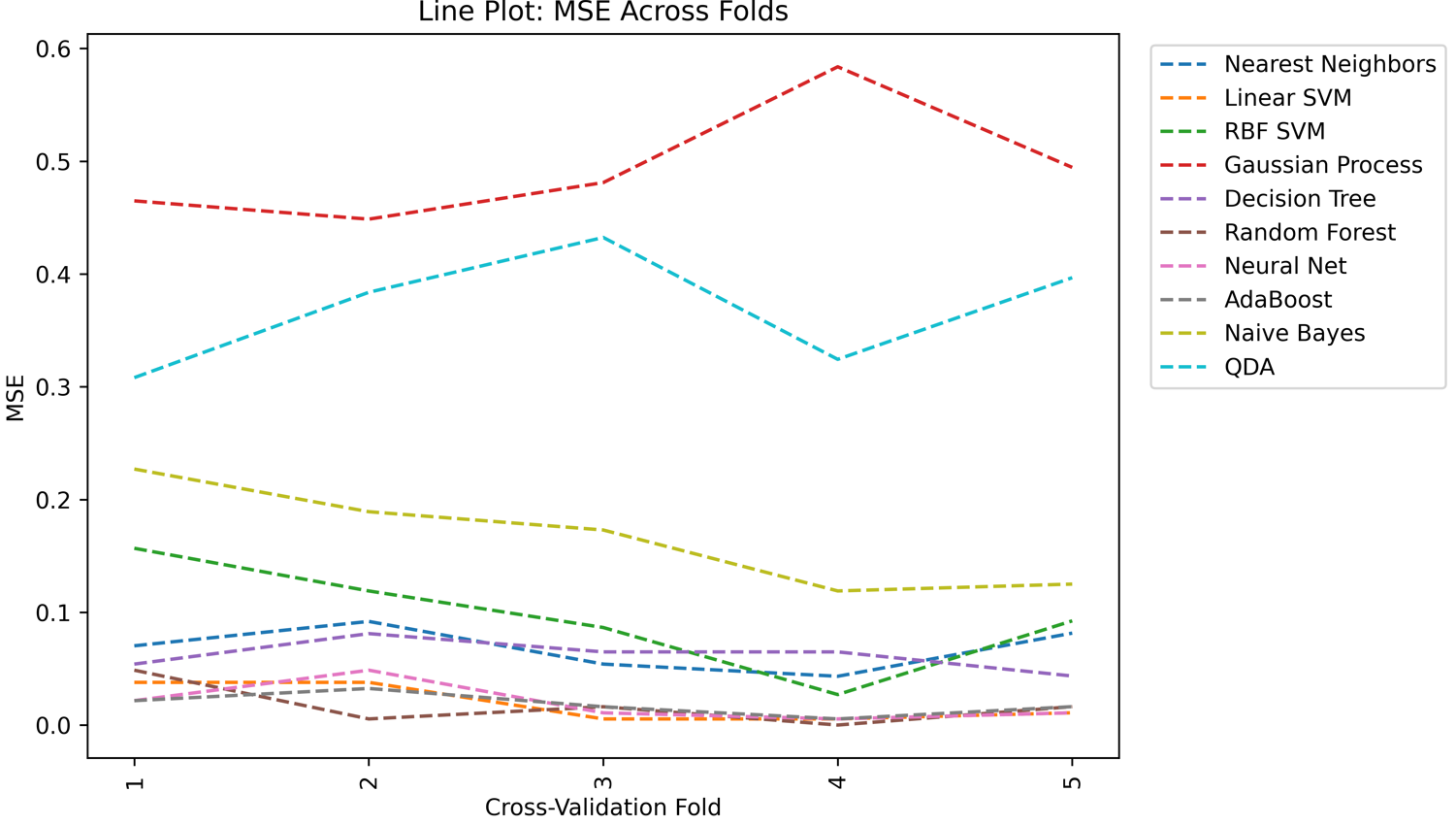}
    \caption{Line plots showing mean square error across five folds for each of the ten classifiers.}
    \label{fig:mse_folds}
\end{figure}

Table~\ref{tab:per_class_metrics} reports precision, recall and F1-score for the two
classes i.e., healthy (H) and inoculated (I) averaged over the five validation folds.
Figure~\ref{fig:cm_grid} visualises the corresponding confusion matrices; the
accuracy and MSE box-and-line plots (described in the text and shown in
Figure~\ref{fig:acc_boxplot},Figure~\ref{fig:acc_folds},Figure~\ref{fig:mse_boxplot},Figure~\ref{fig:mse_folds}).
% Ensure spacing and column formatting
\setlength\tabcolsep{3pt}          
\renewcommand\arraystretch{1.05}   
\newcolumntype{C}{>{\centering\arraybackslash}X}

\begin{table}[H]
\centering
\caption{Per-class classification metrics averaged across 5-fold cross-validation.}
\label{tab:per_class_metrics}
\begin{tabular}{lcccccc}
\hline
\textbf{Classifier} & \multicolumn{3}{c}{\textbf{Healthy (H)}} & \multicolumn{3}{c}{\textbf{Inoculated (I)}} \\
\cmidrule(lr){2-4} \cmidrule(lr){5-7}
 & Precision & Recall & F1-score & Precision & Recall & F1-score \\
\hline
Nearest Neighbors & 0.91 & 0.92 & 0.92 & 0.92 & 0.91 & 0.92 \\
Linear SVM        & 0.98 & 1.00 & 0.99 & 1.00 & 0.98 & 0.99 \\
RBF SVM           & 1.00 & 0.81 & 0.90 & 0.85 & 1.00 & 0.92 \\
Gaussian Process  & 0.50 & 1.00 & 0.67 & 1.00 & 0.02 & 0.04 \\
Decision Tree     & 0.94 & 0.98 & 0.96 & 0.98 & 0.94 & 0.96 \\
Random Forest     & 1.00 & 0.97 & 0.98 & 0.97 & 1.00 & 0.98 \\
Neural Net        & 0.98 & 1.00 & 0.99 & 1.00 & 0.98 & 0.99 \\
AdaBoost          & 0.97 & 1.00 & 0.98 & 1.00 & 0.97 & 0.98 \\
Naive Bayes       & 0.91 & 0.82 & 0.87 & 0.84 & 0.92 & 0.88 \\
QDA               & 0.59 & 0.66 & 0.62 & 0.62 & 0.55 & 0.58 \\
\hline
\end{tabular}
\end{table}

\vspace{4pt}\noindent

Across all the ten different machine learning classifiers, ensemble learners and margin-based models delivered the most reliable discrimination between healthy and SDS-infected leaves. Random Forest, AdaBoost, Neural Net, and Linear SVM consistently exceeded 98\,\% median accuracy, with per-class F1-scores approaching~1.00 and off-diagonal confusion below 3\,\% (Fig.~\ref{fig:cm_grid}\,b,f–h). Such performance indicates that the five GA-selected bands, once passed through the lightweight CNN feature extractor, generate a representation that remains largely linearly separable; consequently, even the compact Linear SVM attains parity with deeper networks.

These results compare favorably to previous SDS detection studies. For instance, Bajwa et al.~\cite{Bajwa2017} utilized hyperspectral reflectance at the leaf level to detect SDS and achieved promising classification rates using linear discriminant analysis and SVM. However, their system lacked real-time integration and involved full-spectrum analysis, whereas our study narrows the input to just five optimal bands using genetic algorithms, resulting in comparable or improved performance while drastically reducing computational overhead. Similarly, Raza et al.~\cite{Raza2020} used PlanetScope satellite imagery to detect SDS at the field scale, applying random forest and SVM on vegetation indices. While their work highlights scalability, it relies on lower-resolution imagery and requires temporal data aggregation. In contrast, our CNN-extracted features operate on high-resolution leaf-level hyperspectral data, offering potential for in-situ decision-making through the accompanying web interface.

A modest gap separates the leading group from the next tier. Nearest-Neighbours and the shallow Decision Tree both maintain mean accuracies above 0.92, yet each shows a mild reduction in recall for infected tissue (Table~\ref{tab:per_class_metrics}); KNN’s distance-based vote is more susceptible to local spectral variability, whereas the Decision Tree occasionally under-generalises, yielding small clusters of misclassified samples visible in Fig.~\ref{fig:cm_grid}\,e.

The remaining classifiers diverge noticeably. RBF-SVM exhibits an asymmetric confusion profile, favouring perfect recall for infected leaves at the expense of healthy classification. Gaussian Process (GP) and Quadratic Discriminant Analysis perform least accurately, with overall accuracies below 65\,\% and correspondingly large mean-squared errors, largely because the GP rarely predicts the infected class whereas QDA shows the opposite bias.

Ensemble models turned out to be superior over others in terms of precision and recall as well. Both Random Forest and AdaBoost achieve precision and recall $\geq 0.97$ for each class, a property shared by the Neural Net and Linear SVM. In contrast, Naive Bayes sacrifices recall on healthy leaves, though its symmetric precision keeps the overall F1-score at a respectable value of 0.88.

Practical deployment considerations strengthen the case for the ensemble and linear models. Random Forest and AdaBoost offer state-of-the-art accuracy and millisecond inference times within the Streamlit front-end web application, while the Linear SVM delivers similar accuracy with an even smaller memory imprint. The Neural Net, although larger, remains attractive when used for multi-class classification (e.g., classification of soybean leaves with multiple diseases) tasks.

The distribution of classification accuracy across five folds for all ten classifiers is shown in Figure~\ref{fig:acc_boxplot}. Ensemble models such as Random Forest and AdaBoost, along with Neural Net and Linear SVM, exhibit consistently high accuracy ($\geq$ 0.98) with minimal variance across folds. Their performance is further confirmed in Figure~\ref{fig:acc_folds}, which plots per-fold accuracy trends and reveals their stability. In contrast, classifiers such as Gaussian Process and QDA display substantial variability and lower overall performance.

Figure~\ref{fig:mse_boxplot} presents the corresponding box plot of mean squared error (MSE), reinforcing the superior precision of the top-tier models. The Neural Net, for instance, maintains an exceptionally low MSE across all folds. This consistency is again evident in Figure~\ref{fig:mse_folds}, where fold-wise MSE trends highlight the robustness of the ensemble models and Linear SVM. Conversely, Gaussian Process yields an unacceptably high MSE due to frequent misclassification of infected samples, while QDA exhibits instability in both directions. 

Collectively, these findings address critical knowledge gaps in early SDS detection and demonstrate a practical, deployable solution that balances performance, speed, and interpretability. In contrast to traditional field-scale SDS risk models like that of Leandro et al.~\cite{Leandro2013}, which rely on environmental parameters to estimate disease pressure, our system provides direct disease state classification from hyperspectral imagery. By enabling real-time disease identification at the leaf scale, our tool complements epidemiological forecasts with actionable phenotypic insights. 

Therefore, combining genetic-algorithm band selection with a lightweight CNN yields a highly discriminative spectral–spatial descriptor. When paired with ensemble or linear classifiers, the approach surpasses 98\,\% accuracy while maintaining computational efficiency compatible with real-time, grower-facing deployment.


\FloatBarrier  % Place just before \section{Conclusion}
\section{Conclusion and Future Work}\label{sec:conclusion}

This study demonstrated a lightweight and accurate pipeline for classifying soybean leaves as healthy or \textit{F.~virguliforme}–infected using hyperspectral imaging data. By integrating a genetic algorithm–based selection of five informative spectral bands centered at 505.4~nm, 563.7~nm, 712.2~nm, 812.9~nm, and 908.4~nm with a shallow CNN for spatial–spectral feature extraction, we enabled efficient and interpretable classification using conventional machine learning models. These bands fall in key visible and near-infrared regions associated with plant physiological traits such as pigment composition and cellular structure, and were found to preserve class separability while reducing input dimensionality.

Among the ten classifiers evaluated, ensemble methods (Random Forest and AdaBoost), as well as Neural Net and Linear SVM, achieved state-of-the-art performance, consistently exceeding 98\% accuracy and maintaining low misclassification errors across five validation folds. The stability of their performance across folds highlights their robustness and suitability for decision-support systems. Conversely, Gaussian Process and QDA classifiers showed inconsistent results, likely due to model assumptions being violated under the spectral data distribution.

Our results also emphasize that compact models like Linear SVM can perform similar to more complex networks when paired with optimized input features, supporting their use in real-time applications. The entire pipeline was deployed in a user friendly web application tool built with Streamlit, allowing growers, extension specialists, plant pathologists and researchers to upload leaf scans and receive instant disease classification and spectral diagnostics.

Our future work will focus on expanding the dataset to include a wider range of soybean genotypes, environmental conditions, and disease stages, including possible abiotic stress confounders. Annotating disease severity levels will enable multiclass classification and monitoring of disease progression. Additionally, we plan to adapt the workflow for field scale deployment under natural illumination by incorporating spectral normalization and domain adaptation techniques. Finally, we aim to generalize this approach to other foliar diseases and crop types, broadening its applicability in plant health monitoring and precision agriculture.

\section*{Acknowledgment}

This research was supported by the South Dakota Soybean Research \& Promotion Council (Grant \#3X5028). The authors gratefully acknowledge the United States Department of Agriculture–Agricultural Research Service (USDA–ARS) and the University of Florida for providing access to the hyperspectral imaging system used in this study. Their technical support and collaboration were instrumental to the successful completion of this work.

\bibliographystyle{IEEEtran}

%\bibliography{references}

\begin{thebibliography}{99}

\bibitem{herrmann2018fusarium}
I. Herrmann, A. Shapiro, L. Nafchi, M. A. Rutter, C. A. Bradley, and S. J. Hill (2018).
Leaf and canopy level detection of \textit{Fusarium virguliforme} (sudden death syndrome) in soybean.
\textit{Remote Sensing}, \textbf{10}(3).

\bibitem{wrather2009effects}
J. A. Wrather and S. R. Koenning (2009).
Effects of diseases on soybean yields in the United States.
\textit{Plant Health Progress}.

\bibitem{cpn2022losses}
Crop Protection Network (2022).
Soybean disease loss estimates from the United States and Ontario, Canada—2022.
\textit{Crop Protection Network Publications}. Retrieved from \url{https://cropprotectionnetwork.org}.

\bibitem{cui2014method}
D. Cui, Q. Zhang, M. Li, T. Slaminko, and G. L. Hartman (2014).
A method for determining the severity of sudden death syndrome in soybeans.
\textit{Transactions of the ASABE}, \textbf{57}(2), 671–678.

\bibitem{bradley2021soybean}
C. A. Bradley et al. (2021).
Soybean yield loss estimates due to diseases in the United States and Ontario, Canada, from 2015 to 2019.
\textit{Plant Health Progress}, \textbf{22}(4), 483–495.

\bibitem{li2003molecular}
S. Li et al. (2003).
Molecular detection of \textit{Fusarium solani} f. sp. \textit{glycines} in soybean roots and soil.
\textit{Plant Pathology}, \textbf{52}(5), 681–689.

\bibitem{cho2001comparison}
H.-Y. Cho and G. L. Hartman (2001).
Comparison of two diagnostic techniques for \textit{Fusarium solani} f. sp. \textit{glycines} in soybean roots.
\textit{Plant Disease}, \textbf{85}(5), 489–493.

\bibitem{frederick2023bands}
Q. Frederick, P. K. Yadav et al. (2023).
Selecting hyperspectral bands and extracting features with a custom shallow CNN to classify citrus peel defects.
\textit{Smart Agricultural Technology}, \textbf{6}, 100365.

\bibitem{yadav2022citrus}
P. K. Yadav et al. (2022).
Citrus disease detection using CNN-generated features and softmax classifier on hyperspectral image data.
\textit{Frontiers in Plant Science}, \textbf{13}, 1043712.

\bibitem{yadav2023spiea}
P. K. Yadav et al. (2023).
Citrus disease classification with CNN features and ML classifiers on hyperspectral image data.
In \textit{Autonomous Air and Ground Sensing Systems for Agricultural Optimization and Phenotyping VIII}, SPIE.

\bibitem{sharma2016ga}
A. Sharma et al. (2016).
Hyperspectral band selection using genetic algorithm and support vector machines for early detection of plant diseases.
\textit{IEEE GRSL}, \textbf{13}(12), 1882–1886.

\bibitem{yin2019hsiband}
X. Yin et al. (2019).
Band selection for hyperspectral imagery classification using weighted neighborhood entropy and genetic algorithm.
\textit{IEEE Access}, \textbf{7}, 177286–177296.

\bibitem{rupe1989fusarium}
J. C. Rupe (1989).
\textit{Fusarium solani} f. sp. \textit{glycines} as a soybean pathogen.
\textit{Plant Disease}, \textbf{73}(7), 581–584.

\bibitem{chawla2013public}
S. Chawla et al. (2013).
A public program to evaluate commercial soybean cultivars for pathogen and pest resistance.
\textit{Plant Disease}, \textbf{97}(5), 568–578.

\bibitem{herman2023evaluation}
T. Herman et al. (2023).
Evaluation of soybean germplasm for resistance to \textit{Fusarium virguliforme}.
\textit{Crop Science}, \textbf{63}(3), 1344–1353.

\bibitem{okello2023fusarium}
P. Okello et al. (2023).
Sources of resistance and marker-trait associations for \textit{Fusarium graminearum} causing root rot in soybean.
\textit{Plant Health Progress}, \textbf{24}(4), 411–415.

\bibitem{iastate_growth}
Iowa State University Extension (2023).
How to identify growth stages. Accessed: 30 June 2025.
\url{https://crops.extension.iastate.edu/encyclopedia/soybean-growth-stages}.

\bibitem{frederick2025supervised}
Q. Frederick et al. (2025).
Supervised hyperspectral band selection using texture features for citrus leaf disease classification with YOLOv8.
\textit{Sensors}, \textbf{25}(4), 1034.

\bibitem{goldberg1989genetic}
D. E. Goldberg (1989).
\textit{Genetic Algorithms in Search, Optimization and Machine Learning}.
Addison-Wesley.

\bibitem{nagasubramanian2018hyperspectral}
K. Nagasubramanian et al. (2018).
Genetic algorithm and SVM-based band selection for early identification of charcoal rot in soybean stems.
\textit{Plant Methods}, \textbf{14}, 1–13.

\bibitem{DEAP_JMLR2012}
F.-A. Fortin et al. (2012).
DEAP: Evolutionary algorithms made easy.
\textit{Journal of Machine Learning Research}, \textbf{13}, 2171–2175.

\bibitem{streamlit2023}
Streamlit Inc. (2023).
Streamlit: A faster way to build and share data apps.
\url{https://github.com/streamlit/streamlit}. Accessed: 2025-07-02.

\bibitem{lie2005css}
H. W. Lie and B. Bos (2005).
\textit{Cascading Style Sheets: Designing for the Web} (3rd ed.).
Addison-Wesley Professional.

\bibitem{wolf2017css}
D. Wolf and A. J. Henley (2017).
Cascading style sheets (CSS).
In \textit{Java EE Web Application Primer}, Apress, pp. 115–118.

\bibitem{Bajwa2017}
S. G. Bajwa et al. (2017).
Hyperspectral reflectance sensing for detection of soybean sudden death syndrome.
\textit{Precision Agriculture}, \textbf{18}(2), 249–261.

\bibitem{Raza2020}
S. A. Raza et al. (2020).
Soybean sudden death syndrome detection at field scale using PlanetScope imagery.
\textit{Remote Sensing}, \textbf{12}(7), 1213.

\bibitem{Leandro2013}
L. F. Leandro et al. (2013).
Predicting severity of soybean SDS and associated yield loss in North-Central U.S.
\textit{Plant Health Progress}, \textbf{14}(2), 45–53.

\end{thebibliography}
\end{document}